\documentclass{article}
\usepackage{spconf,amsmath,epsfig}
\usepackage{times}
\usepackage{epsfig}
\usepackage{graphicx}
\usepackage{amsmath}
\usepackage{amssymb}
\usepackage{booktabs}
\usepackage{multirow}
\usepackage{verbatim}
\usepackage{float}
\usepackage{subfigure} 
\usepackage[preprint]{spconfa4}

\copyrightnotice{978-1-6654-3864-3/21/\$31.00~\copyright 2021 IEEE}

\let\OLDthebibliography\thebibliography
\renewcommand\thebibliography[1]{
  \OLDthebibliography{#1}
  \setlength{\parskip}{0pt}
  \setlength{\itemsep}{0pt plus 0.3ex}
}

\begin{document}\sloppy

% Example definitions.
% --------------------
\def\x{{\mathbf x}}
\def\L{{\cal L}}

% Title.
% ------
\title{A Dataset And Benchmark Of Underwater Object Detection for Robot Picking}
%
% Address.
% ---------------

\name{Chongwei Liu, Haojie Li, Shuchang Wang, Ming Zhu, Dong Wang, Xin Fan, and Zhihui Wang$^{*}$ 
}

\address{
DUT-RU International School of Information Science \& Engineering, Dalian University of Technology\\
\{lcwdllg, lwiiy\}@mail.dlut.edu.cn\\
\{hjli, zhuming, wdice, zhwang\}@dlut.edu.cn}
\maketitle

\maketitle
\begin{abstract}
Underwater object detection for robot picking has attracted a lot of interest. However, it is still an unsolved problem due to several challenges. We take steps towards making it more realistic by addressing the following challenges. Firstly, the currently available datasets basically lack the test set annotations, causing researchers must compare their method with other \emph{SOTAs} on a self-divided test set (from the training set). Training other methods lead to an increase in workload and different researchers divide different datasets, resulting there is no unified benchmark to compare the performance of different algorithms. Secondly, these datasets also have other shortcomings, \emph{e.g.}, too many similar images or incomplete labels. Towards these challenges we introduce a dataset, Detecting Underwater Objects (DUO), and a corresponding benchmark, based on the collection and re-annotation of all relevant datasets. DUO contains a collection of diverse underwater images with more rational annotations. The corresponding benchmark provides indicators of both efficiency and accuracy of \emph{SOTAs} (under the MMDtection framework) for academic research and industrial applications, where JETSON AGX XAVIER is used to assess detector speed to simulate the robot-embedded environment.
\end{abstract}
\begin{keywords}
Underwater object detection, robot picking, dataset, benchmark
\end{keywords}
\section{Introduction}
Underwater robot picking is to use the robot to automatically capture sea creatures like holothurian, echinus, scallop, or starfish in an open-sea farm where underwater object detection is the key technology for locating creatures. Until now, the datasets used in this community are released by the Underwater Robot Professional Contest (URPC$\protect\footnote{Underwater Robot Professional Contest: {\bf http://en.cnurpc.org}.}$) beginning from 2017, in which URPC2017 and URPC2018 are most often used for research. Unfortunately, as the information listed in Table \ref{Info}, URPC series datasets do not provide the annotation file of the test set and cannot be downloaded after the contest. 
Therefore, researchers \cite{2020arXiv200511552C,2019arXiv191103029L} first have to divide the training data into two subsets, including a new subset of training data and a new subset of testing data, and then train their proposed method and other \emph{SOTA} methods. On the one hand, training other methods results in a significant increase in workload. On the other hand, different researchers divide different datasets in different ways, 
\begin{table}[t]
\renewcommand\tabcolsep{3.5pt}
\caption{Information about all the collected datasets. * denotes the test set's annotations are not available. \emph{3} in Class means three types of creatures are labeled, \emph{i.e.,} holothurian, echinus, and scallop. \emph{4} means four types of creatures are labeled (starfish added). Retention represents the proportion of images that retain after similar images have been removed.}
\centering 
\begin{tabular}{|l|c|c|c|c|c|}
\hline
Dataset&Train&Test&Class&Retention&Year \\ 
\hline  % 中部线
URPC2017&17,655&985*&3&15\%&2017 \\
\hline
URPC2018&2,901&800*&4&99\%&2018 \\
\hline
URPC2019&4,757&1,029*&4&86\%&2019 \\
\hline
URPC2020$_{ZJ}$&5,543&2,000*&4&82\%&2020 \\
\hline
URPC2020$_{DL}$&6,575&2,400*&4&80\%&2020 \\
\hline
UDD&1,827&400&3&84\%&2020 \\
\hline  % 底部线

\end{tabular}
\label{Info}
\end{table}
\begin{figure*}[htbp]
\begin{center}
\includegraphics[width=1\linewidth]{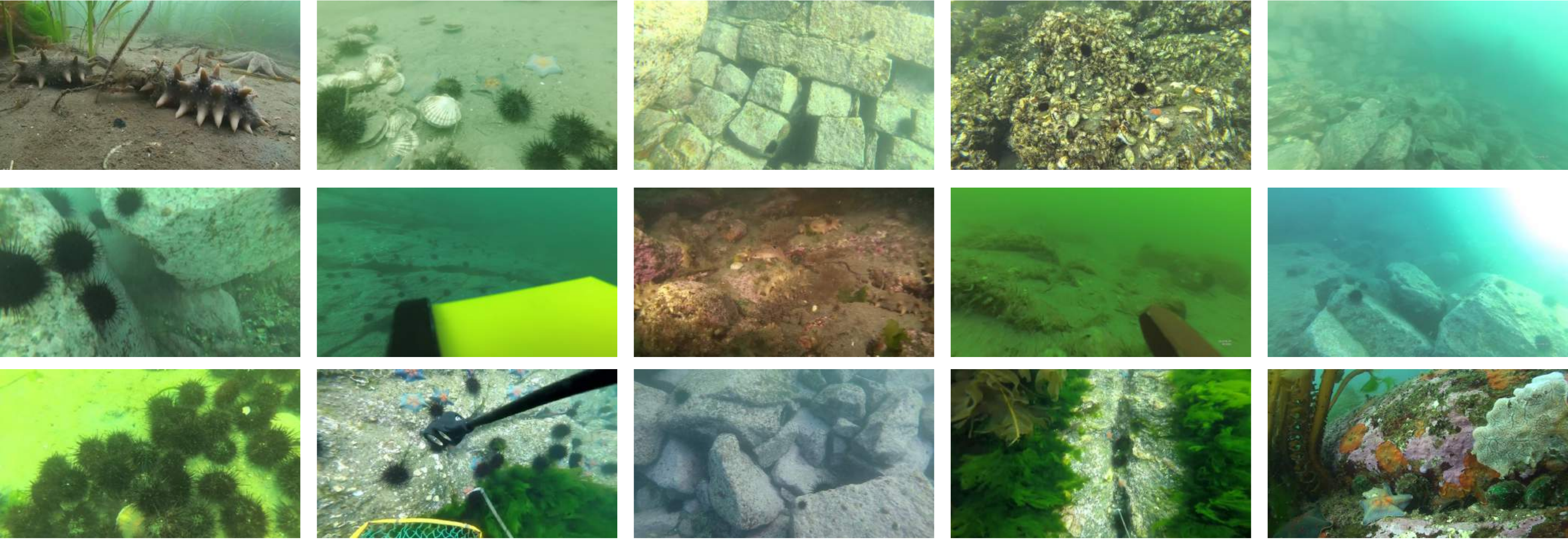}
\end{center}
   \caption{Examples in DUO, which show a variety of scenarios in underwater environments.}
\label{exam}
\end{figure*}
causing there is no unified benchmark to compare the performance of different algorithms.
In terms of the content of the dataset images, there are a large number of similar or duplicate images in the URPC datasets. URPC2017 only retains 15\% images after removing similar images compared to other datasets. Thus the detector trained on URPC2017 is easy to overfit and cannot reflect the real performance.
% For URPC2017, only 15\% images remain after deleting overly similar images 
% because it was generated from several videos at an interval of 10 frames,
%  which means detectors trained on it may overfit and could not reflect true accuracy. Because the frequency of underwater robot capture scene changes is low, the content change of the sample picture taken only 10 frames apart is usually very small. CITE 
For other URPC datasets, the latter also includes images from the former, \emph{e.g.}, URPC2019 adds 2,000 new images compared to URPC2018; compared with URPC2019, URPC2020$_{ZJ}$ adds 800 new images. The URPC2020$_{DL}$ adds 1,000 new images compared to the URPC2020$_{ZJ}$. It is worth mentioning that the annotation of all datasets is incomplete; some datasets lack the starfish labels and it is easy to find error or missing labels. \cite{DBLP:conf/iclr/ZhangBHRV17} pointed out that although the CNN model has a strong fitting ability for any dataset, the existence of dirty data will significantly weaken its robustness.
Therefore, a reasonable dataset (containing a small number of similar images as well as an accurate annotation) and a corresponding recognized benchmark are urgently needed to promote community development.

To address these issues, we introduce a dataset called Detecting Underwater Objects (DUO) by collecting and re-annotating all the available underwater datasets. It contains 7,782 underwater images after deleting overly similar images and has a more accurate annotation with four types of classes (\emph{i.e.,} holothurian, echinus, scallop, and starfish). 
Besides, based on the MMDetection$\protect\footnote{MMDetection is an open source object detection toolbox based on PyTorch. {\bf https://github.com/open-mmlab/mmdetection}}$ \cite{chen2019mmdetection} framework, we also provide a \emph{SOTA} detector benchmark containing efficiency and accuracy indicators, providing a reference for both academic research and industrial applications. It is worth noting that JETSON AGX XAVIER$\protect\footnote{JETSON AGX XAVIER is an embedded development board produced by NVIDIA which could be deployed in an underwater robot. Please refer {\bf https://developer.nvidia.com/embedded/jetson-agx-xavier-developer-kit} for more information.}$ was used to assess all the detectors in the efficiency test in order to simulate robot-embedded environment. DUO will be released in https://github.com/chongweiliu soon.

In summary, the contributions of this paper can be listed as follows.

  $\bullet$ By collecting and re-annotating all relevant datasets, we introduce a dataset called DUO with more reasonable annotations as well as a variety of underwater scenes.

  $\bullet$ We provide a corresponding benchmark of \emph{SOTA} detectors on DUO including efficiency and accuracy indicators which could be a reference for both academic research and industrial applications.

\pagestyle{empty}
\section{Background}
In the year of 2017, underwater object detection for open-sea farming is first proposed in the target recognition track of Underwater Robot Picking Contest 2017$\protect\footnote{From 2020, the name has been changed into Underwater Robot Professional Contest which is also short for URPC.}$ (URPC2017) which aims to promote the development of theory, technology, and industry of the underwater agile robot and fill the blank of the grabbing task of the underwater agile robot. The competition sets up a target recognition track, a fixed-point grasping track, and an autonomous grasping track. The target recognition track concentrates on finding the {\bf high accuracy and efficiency} algorithm which could be used in an underwater robot for automatically grasping.

The datasets we used to generate the DUO are listed below. The detailed information has been shown in Table \ref{Info}.

  {\bf URPC2017}: It contains 17,655 images for training and 985 images for testing and the resolution of all the images is 720$\times$405. All the images are taken from 6 videos at an interval of 10 frames. However, all the videos were filmed in an artificial simulated environment and pictures from the same video look almost identical. 
  
   {\bf URPC2018}: It contains 2,901 images for training and 800 images for testing and the resolutions of the images are 586$\times$480, 704$\times$576, 720$\times$405, and 1,920$\times$1,080. The test set's annotations are not available. Besides, some images were also collected from an artificial underwater environment.
  
  {\bf URPC2019}: It contains 4,757 images for training and 1029 images for testing and the highest resolution of the images is 3,840$\times$2,160 captured by a GOPro camera. The test set's annotations are also not available and it contains images from the former contests.
  
  {\bf URPC2020$_{ZJ}$}: From 2020, the URPC will be held twice a year. It was held first in Zhanjiang, China, in April and then in Dalian, China, in August. URPC2020$_{ZJ}$ means the dataset released in the first URPC2020 and URPC2020$_{DL}$ means the dataset released in the second URPC2020. This dataset contains 5,543 images for training and 2,000 images for testing and the highest resolution of the images is 3,840$\times$2,160. The test set's annotations are also not available.
  
  {\bf URPC2020$_{DL}$}: This dataset contains 6,575 images for training and 2,400 images for testing and the highest resolution of the images is 3,840$\times$2,160. The test set's annotations are also not available.
  
  {\bf UDD \cite{2020arXiv200301446W}}: This dataset contains 1,827 images for training and 400 images for testing and the highest resolution of the images is 3,840$\times$2,160. All the images are captured by a diver and a robot in a real open-sea farm.

\begin{figure}[t]
\begin{center}
\includegraphics[width=1\linewidth]{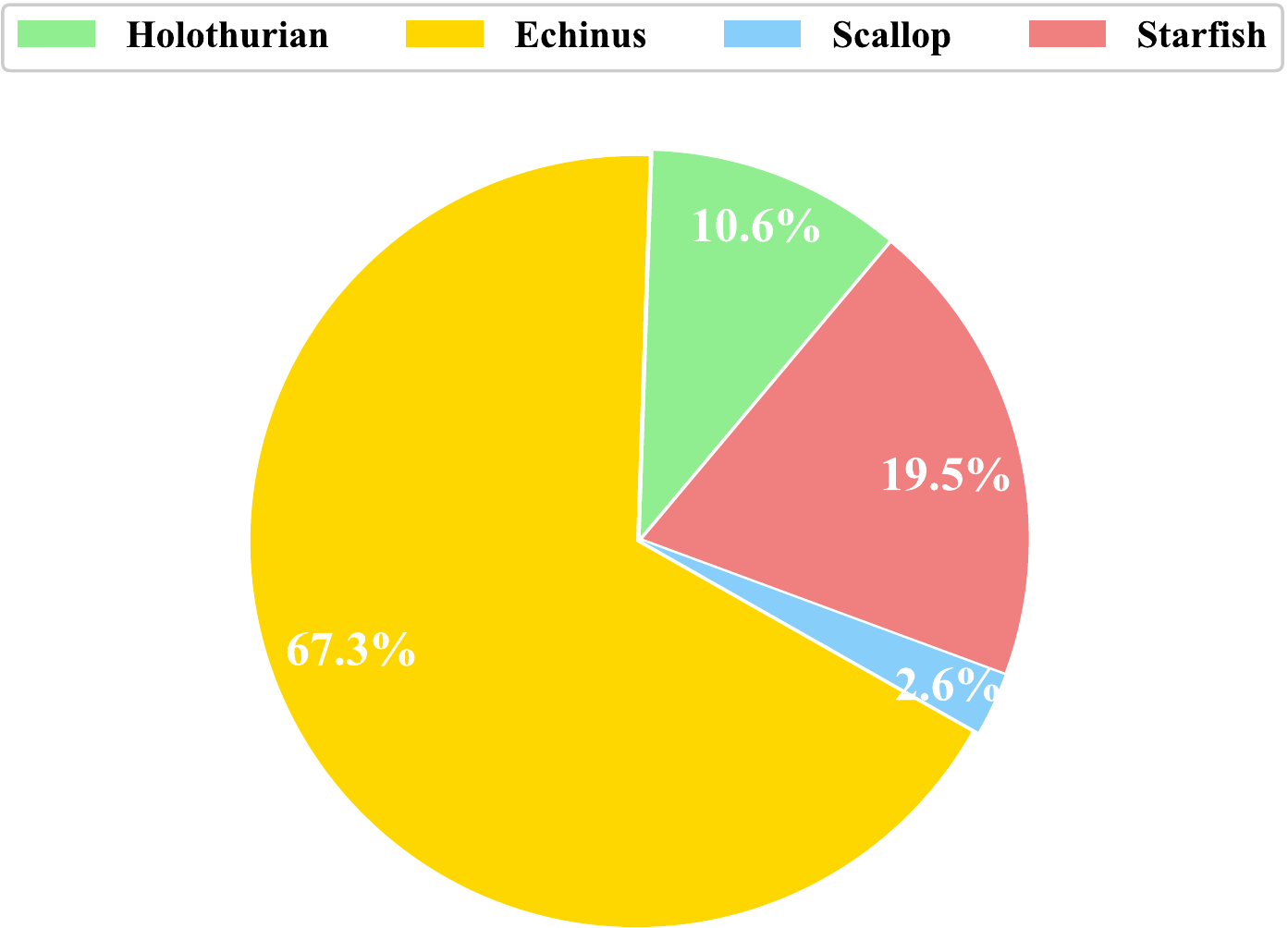}
\end{center}
   \caption{The proportion distribution of the objects in DUO.}
\label{pie}
\end{figure}
% \subsection{Analysis of the Datasets}
% First, the UPRC series datasets do not provide the download URLs after finishing and also do not provide the annotation files of testing sets, making researchers divide the training set into a new training set and a new testing set to train their proposed and other \emph{SOTA} methods and then compare them. 

% About the image, too many similar images appear in one dataset, \emph{e.g.}, only 15.7\% of the images in URPC2017 are left after deleting overly similar images, because all the datasets were generated from several videos at an interval of 10 frames. In addition, for URPC series datasets, the later one may also include some images from the former one (URPC2019 has all the images in URPC2018).

% Besides, the annotations of all the datasets are not complete; error labels and miss labels are easy to find in these datasets. 
% because it is very difficult for a person to annotate the correct label from blur underwater scenarios.

\begin{figure*}
  \centering
  \subfigure[]{\includegraphics[width=3.45in]{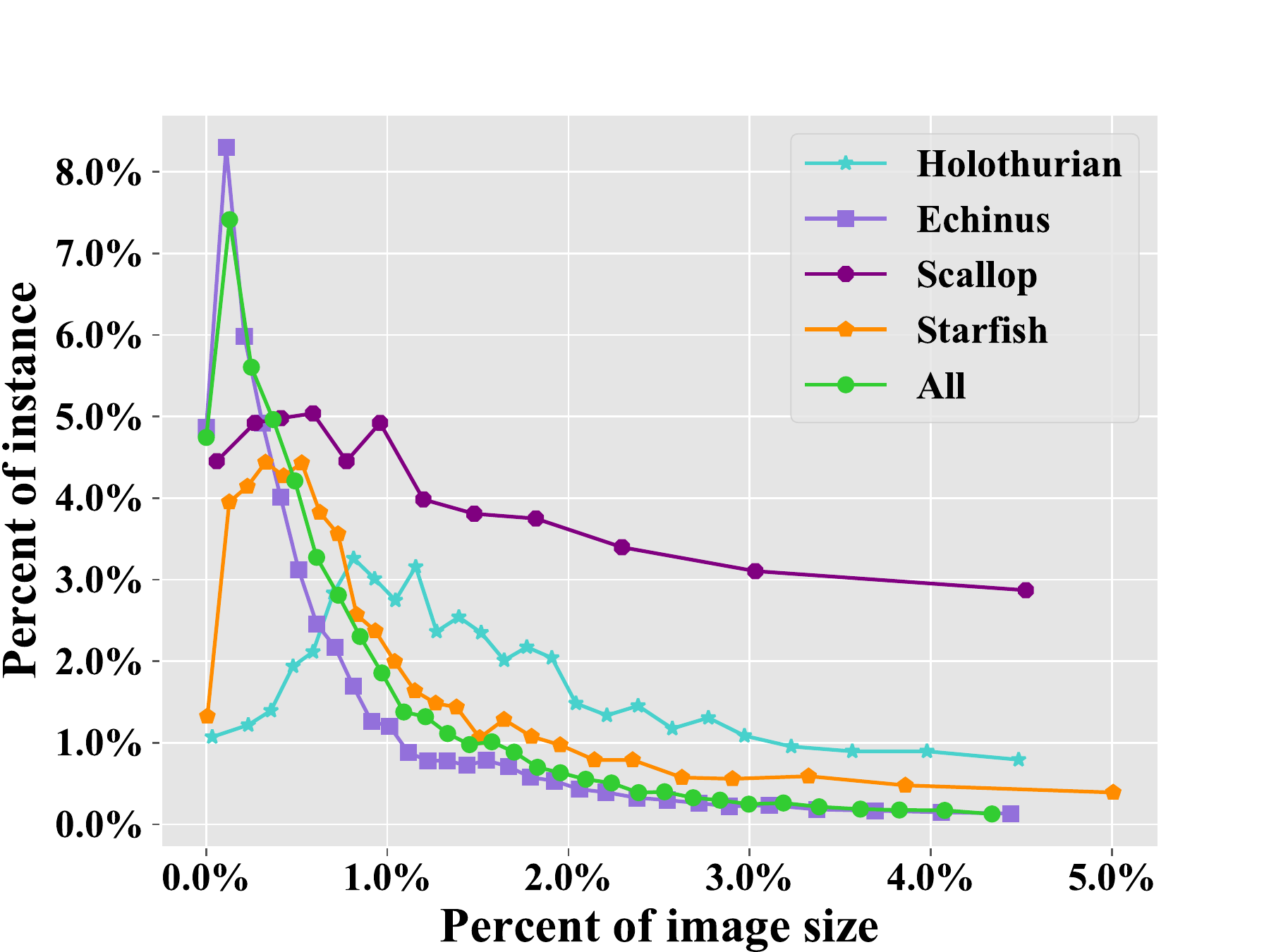}}
  \subfigure[]{\includegraphics[width=3.45in]{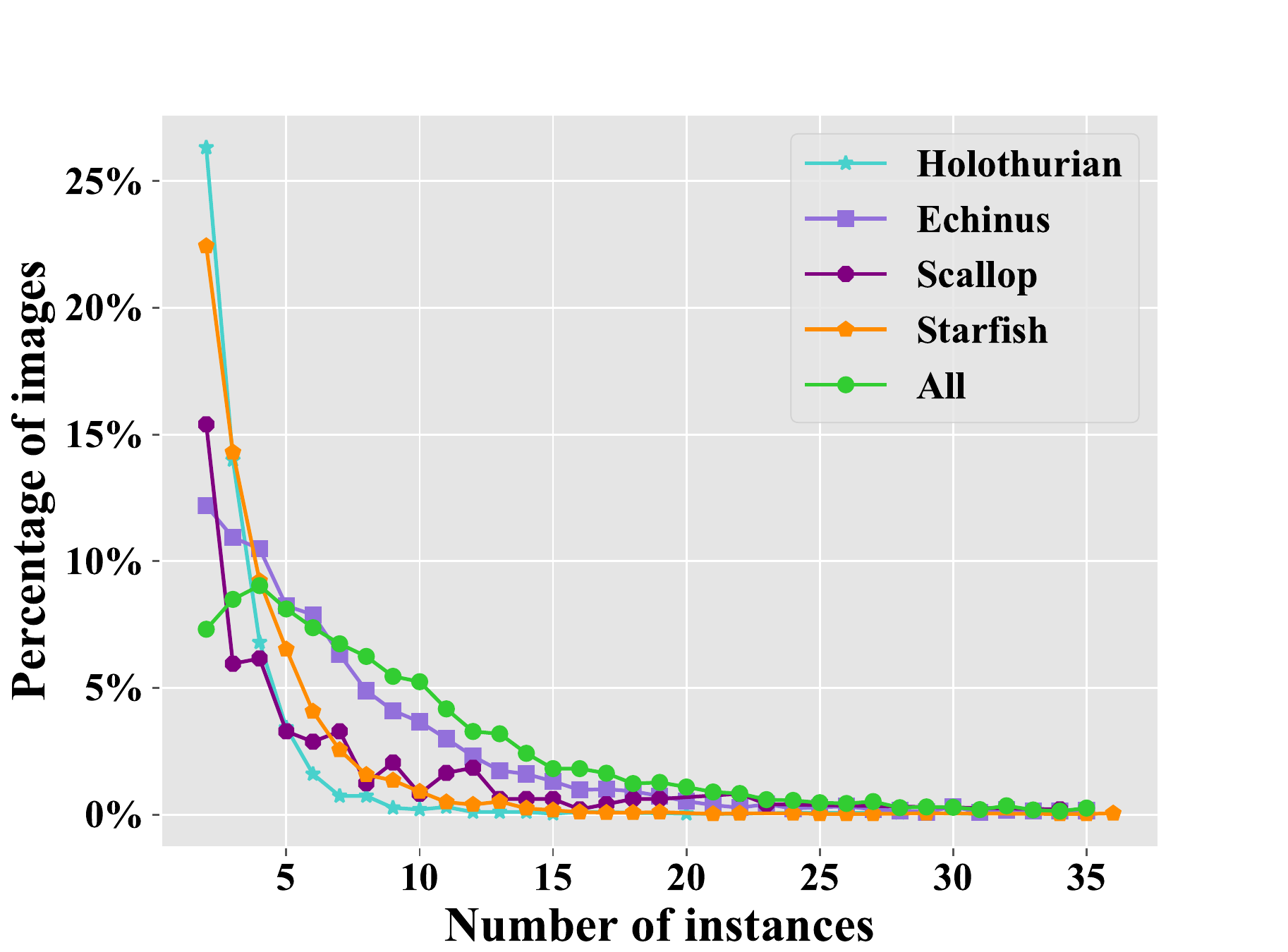}}
  \caption{(a) The distribution of instance sizes for DUO; (b) The number of categories per image.}
  \label{sum}
\end{figure*}
\section{Proposed Dataset}

% In this section, we will describe the details of our proposed dataset.
\subsection{Image Deduplicating}
% By observing the collected datasets introduced in Section 2.1, we find that some images are very similar within one dataset because the dataset was generated from the same videos. In URPC series datasets, the later one (\emph{e.g.}, URPC2020$_{ZJ}$) has some images from the former ones (\emph{e.g.}, URPC2017 or URPC2018). 
As we explained in Section 1, there are a large number of similar or repeated images in the series of URPC datasets. Therefore, it is important to delete duplicate or overly similar images and keep a variety of underwater scenarios when we merge these datasets together. Here we employ the Perceptual Hash algorithm (PHash) to remove those images. PHash has the special property that the hash value is dependent on the image content, and it remains approximately the same if the content is not significantly modified. Thus we can easily distinguish different scenarios and delete duplicate images within one scenario. 

After deduplicating, we obtain 7,782 images (6,671 images for training; 1,111 for testing). The retention rate of the new dataset is 95\%, which means that there are only a few similar images in the new dataset. Figure \ref{exam} shows that our dataset also retains various underwater scenes.

\subsection{Image Re-annotation}
Due to the small size of objects and the blur underwater environment, there are always missing or wrong labels in the existing annotation files. In addition, some test sets' annotation files are not available and some datasets do not have the starfish annotation. In order to address these issues, we follow the next process which combines a CNN model and manual annotation to re-annotate these images. Specifically, we first train a detector (\emph{i.e.,} GFL \cite{li2020generalized}) with the originally labeled images. After that, the trained detector predicts all the 7,782 images. We treat the prediction as the groundtruth and use it to train the GFL again. We get the final GFL prediction called {\bf the coarse annotation}. Next, we use manual correction to get the final annotation called {\bf the fine annotation}. Notably, we adopt the COCO \cite{Belongie2014} annotation form as the final format.
% \subsection{Split}
% Eventually, we obtain the new dataset, called DUO, which has 7,887 images totally. After that, we choose one picture every seven in the total sequence to generate the testing set. The training set includes 6,671 images and the testing set includes 1,111 images.
\subsection{Dataset Statistics}
{\bf The proportion of classes}: The total number of objects is 74,515. Holothurian, echinus, scallop, and starfish are 7,887, 50,156, 1,924, and 14,548, respectively. Figure \ref{pie} shows the proportion of each creatures where echinus accounts for 67.3\% of the total. The whole data distribution shows an obvious long-tail distribution because the different economic benefits of different seafoods determine the different breed quantities.

{\bf The distribution of instance sizes}: Figure \ref{sum}(a) shows an instance size distribution of DUO. \emph{Percent of image size} represents the ratio of object area to image area, and \emph{Percent of instance} represents the ratio of the corresponding number of objects to the total number of objects. Because of these small creatures and high-resolution images, the vast majority of objects occupy 0.3\% to 1.5\% of the image area.

{\bf The instance number per image}: Figure \ref{sum}(b) illustrates the number of categories per image for DUO. \emph{Number of instances} represents the number of objects one image has, and \emph{ Percentage of images} represents the ratio of the corresponding number of images to the total number of images. Most images contain between 5 and 15 instances, with an average of 9.57 instances per image.

{\bf Summary}:
In general, smaller objects are harder to detect. For PASCAL VOC \cite{Everingham2007The} or COCO \cite{Belongie2014}, roughly 50\% of all objects occupy no more than 10\% of the image itself, and others evenly occupy from 10\% to 100\%. 
In the aspect of instances number per image, COCO contains 7.7 instances per image and VOC contains 3. In comparison, DUO has 9.57 instances per image and most instances less than 1.5\% of the image size.
Therefore, DUO contains almost exclusively massive small instances and has the long-tail distribution at the same time, which means it is promising to design a detector to deal with massive small objects and stay high efficiency at the same time for underwater robot picking.

\section{Benchmark}
Because the aim of underwater object detection for robot picking is to find {\bf the high accuracy and efficiency} algorithm, we consider both the accuracy and efficiency evaluations in the benchmark as shown in Table \ref{ben}.

\subsection{Evaluation Metrics}
Here we adopt the standard COCO metrics (mean average precision, \emph{i.e.,} mAP) for the accuracy evaluation and also provide the mAP of each class due to the long-tail distribution.

{\bf AP} -- mAP at IoU=0.50:0.05:0.95.

{\bf AP$_{50}$} --  mAP at IoU=0.50.

{\bf AP$_{75}$} --  mAP at IoU=0.75. 

{\bf AP$_{S}$} --   {\bf AP} for small objects of area smaller than 32$^{2}$.

{\bf AP$_{M}$} --   {\bf AP} for objects of area between 32$^{2}$ and 96$^{2}$.

{\bf AP$_{L}$} --   {\bf AP} for large objects of area bigger than 96$^{2}$.

{\bf AP$_{Ho}$} --  {\bf AP} in holothurian.

{\bf AP$_{Ec}$} --  {\bf AP} in echinus.

{\bf AP$_{Sc}$} --  {\bf AP} in scallop.

{\bf AP$_{St}$} --  {\bf AP} in starfish.

% mAP at IoU=0.50:0.05:0.95 in starfish.

For the efficiency evaluation, we provide three metrics:

{\bf Param.} --  The parameters of a detector.

{\bf FLOPs} --  Floating-point operations per second.

{\bf FPS} --  Frames per second.

Notably, {\bf FLOPs} is calculated under the 512$\times$512 input image size and {\bf FPS} is tested on a JETSON AGX XAVIER under MODE$\_$30W$\_$ALL. 

\subsection{Standard Training Configuration}
We follow a widely used open-source toolbox, \emph{i.e.,} MMDetection (V2.5.0) to produce up our benchmark. During the training, the standard configurations are as follows:

% \begin{itemize}
  $\bullet$ We initialize the backbone models (\emph{e.g.,} ResNet50) with pre-trained parameters on ImageNet \cite{Deng2009ImageNet}.

  $\bullet$ We resize each image into 512 $\times$ 512 pixels both in training and testing. Each image is flipped horizontally with 0.5 probability during training.

  $\bullet$ We normalize RGB channels by subtracting 123.675, 116.28, 103.53 and dividing by 58.395, 57.12, 57.375, respectively.

  $\bullet$ SGD method is adopted to optimize the model. The initial learning rate is set to be 0.005 in a single GTX 1080Ti with batchsize 4 and is decreased by 0.1 at the 8th and 11th epoch, respectively. WarmUp \cite{2019arXiv190307071L} is also employed in the first 500 iterations. Totally there are 12 training epochs.

  $\bullet$ Testing time augmentation (\emph{i.e.,} flipping test or multi-scale testing) is not employed.
% \end{itemize}

\subsection{Benchmark Analysis}
Table \ref{ben} shows the benchmark for the \emph{SOTA} methods. Multi- and one- stage detectors with three kinds of backbones (\emph{i.e.,} ResNet18, 50, 101) give a comprehensive assessment on DUO. We also deploy all the methods to AGX to assess efficiency.

In general, the multi-stage (Cascade R-CNN) detectors have high accuracy and low efficiency, while the one-stage (RetinaNet) detectors have low accuracy and high efficiency. However, due to recent studies \cite{zhang2019bridging} on the allocation of more reasonable positive and negative samples in training, one-stage detectors (ATSS or GFL) can achieve both high accuracy and high efficiency.

\begin{table*}[htbp]
\renewcommand\tabcolsep{3.0pt}

\begin{center}
\caption{Benchmark of \emph{SOTA} detectors (single-model and single-scale results) on DUO. FPS is measured on the same machine with a JETSON AGX XAVIER under the same MMDetection framework, using a batch size of 1 whenever possible. R: ResNet.} 
\label{ben}
\begin{tabular}{|l|l|c|c|c|ccc|ccc|cccc|}
\hline
Method&Backbone&Param.&FLOPs&FPS&AP&AP$_{50}$&AP$_{75}$&AP$_{S}$&AP$_{M}$&AP$_{L}$&AP$_{Ho}$&AP$_{Ec}$&AP$_{Sc}$&AP$_{St}$ \\ 
\hline  % 中部线
\emph{multi-stage:} &&&&&&&&&&&&&& \\

\multirow{3}{*}{Faster R-CNN \cite{Ren2015Faster}}
&R-18&28.14M&49.75G&5.7&50.1&72.6&57.8&42.9&51.9&48.7&49.1&60.1&31.6&59.7\\
&R-50&41.14M&63.26G&4.7&54.8&75.9&63.1&53.0&56.2&53.8&55.5&62.4&38.7&62.5\\
&R-101&60.13M&82.74G&3.7&53.8&75.4&61.6&39.0&55.2&52.8&54.3&62.0&38.5&60.4\\
\hline

\multirow{3}{*}{Cascade R-CNN \cite{Cai_2019}}
&R-18&55.93M&77.54G&3.4&52.7&73.4&60.3&\bf 49.0&54.7&50.9&51.4&62.3&34.9&62.3\\
&R-50&68.94M&91.06G&3.0&55.6&75.5&63.8&44.9&57.4&54.4&56.8&63.6&38.7&63.5\\
&R-101&87.93M&110.53G&2.6&56.0&76.1&63.6&51.2&57.5&54.7&56.2&63.9&41.3&62.6\\
\hline

\multirow{3}{*}{Grid R-CNN \cite{lu2019grid}}
&R-18&51.24M&163.15G&3.9&51.9&72.1&59.2&40.4&54.2&50.1&50.7&61.8&33.3&61.9\\
&R-50&64.24M&176.67G&3.4&55.9&75.8&64.3&40.9&57.5&54.8&56.7&62.9&39.5&64.4\\
&R-101&83.24M&196.14G&2.8&55.6&75.6&62.9&45.6&57.1&54.5&55.5&62.9&41.0&62.9\\
\hline

\multirow{3}{*}{RepPoints \cite{yang2019reppoints}}
&R-18&20.11M&\bf 35.60G&5.6&51.7&76.9&57.8&43.8&54.0&49.7&50.8&63.3&33.6&59.2\\
&R-50&36.60M&48.54G&4.8&56.0&80.2&63.1&40.8&58.5&53.7&56.7&65.7&39.3&62.3\\
&R-101&55.60M&68.02G&3.8&55.4&79.0&62.6&42.2&57.3&53.9&56.0&65.8&39.0&60.9\\
\hline  % 中部线
\hline  % 中部线
\emph{one-stage:} &&&&&&&&&&&&&& \\
\multirow{3}{*}{RetinaNet \cite{Lin2017Focal}}
&R-18&19.68M&39.68G&7.1&44.7&66.3&50.7&29.3&47.6&42.5&46.9&54.2&23.9&53.8\\
&R-50&36.17M&52.62G&5.9&49.3&70.3&55.4&36.5&51.9&47.6&54.4&56.6&27.8&58.3\\
&R-101&55.16M&72.10G&4.5&50.4&71.7&57.3&34.6&52.8&49.0&54.6&57.0&33.7&56.3\\
\hline  % 中部线

\multirow{3}{*}{FreeAnchor \cite{2019arXiv190902466Z}}
&R-18&19.68M&39.68G&6.8&49.0&71.9&55.3&38.6&51.7&46.7&47.2&62.8&28.6&57.6\\
&R-50&36.17M&52.62G&5.8&54.4&76.6&62.5&38.1&55.7&53.4&55.3&65.2&35.3&61.8\\
&R-101&55.16M&72.10G&4.4&54.6&76.9&62.9&36.5&56.5&52.9&54.0&65.1&38.4&60.7\\
\hline  % 中部线

\multirow{3}{*}{FoveaBox \cite{DBLP:journals/corr/abs-1904-03797}}
&R-18&21.20M&44.75G&6.7&51.6&74.9&57.4&40.0&53.6&49.8&51.0&61.9&34.6&59.1\\
&R-50&37.69M&57.69G&5.5&55.3&77.8&62.3&44.7&57.4&53.4&57.9&64.2&36.4&62.8\\
&R-101&56.68M&77.16G&4.2&54.7&77.3&62.3&37.7&57.1&52.4&55.3&63.6&38.9&60.8\\
\hline  % 中部线

\multirow{3}{*}{PAA \cite{2020arXiv200708103K}}
&R-18&\bf 18.94M&38.84G&3.0&52.6&75.3&58.8&41.3&55.1&50.2&49.9&64.6&35.6&60.5\\
&R-50&31.89M&51.55G&2.9&56.8&79.0&63.8&38.9&58.9&54.9&56.5&66.9&39.9&64.0\\
&R-101&50.89M&71.03G&2.4&56.5&78.5&63.7&40.9&58.7&54.5&55.8&66.5&42.0&61.6\\
\hline  % 中部线

\multirow{3}{*}{FSAF \cite{zhu2019feature}}
&R-18&19.53M&38.88G&\bf 7.4&49.6&74.3&55.1&43.4&51.8&47.5&45.5&63.5&30.3&58.9\\
&R-50&36.02M&51.82G&6.0&54.9&79.3&62.1&46.2&56.7&53.3&53.7&66.4&36.8&62.5\\
&R-101&55.01M&55.01G&4.5&54.6&78.7&61.9&46.0&57.1&52.2&53.0&66.3&38.2&61.1\\
\hline  % 中部线

\multirow{3}{*}{FCOS \cite{DBLP:journals/corr/abs-1904-01355}}
&R-18&\bf 18.94M&38.84G&6.5&48.4&72.8&53.7&30.7&50.9&46.3&46.5&61.5&29.1&56.6\\
&R-50&31.84M&50.34G&5.4&53.0&77.1&59.9&39.7&55.6&50.5&52.3&64.5&35.2&60.0\\
&R-101&50.78M&69.81G&4.2&53.2&77.3&60.1&43.4&55.4&51.2&51.7&64.1&38.5&58.5\\
\hline  % 中部线

\multirow{3}{*}{ATSS \cite{zhang2019bridging}}
&R-18&\bf 18.94M&38.84G&6.0&54.0&76.5&60.9&44.1&56.6&51.4&52.6&65.5&35.8&61.9\\
&R-50&31.89M&51.55G&5.2&58.2&\bf 80.1&66.5&43.9&60.6&55.9&\bf 58.6&67.6&41.8&64.6\\
&R-101&50.89M&71.03G&3.8&57.6&79.4&65.3&46.5&60.3&55.0&57.7&67.2&42.6&62.9\\
\hline  % 中部线

\multirow{3}{*}{GFL \cite{li2020generalized}}
&R-18&19.09M&39.63G&6.3&54.4&75.5&61.9&35.0&57.1&51.8&51.8&66.9&36.5&62.5\\
&R-50&32.04M&52.35G&5.5&\bf 58.6&79.3&\bf 66.7&46.5&\bf 61.6&55.6&\bf 58.6&\bf 69.1&41.3&\bf 65.3\\
&R-101&51.03M&71.82G&4.1&58.3&79.3&65.5&45.1&60.5&\bf 56.3&57.0&\bf 69.1&\bf 43.0&64.0\\

\hline  % 底部线
\end{tabular}
\end{center}
\end{table*}
Therefore, in terms of accuracy, the accuracy difference between the multi- and the one- stage methods in AP is not obvious, and the AP$_{S}$ of different methods is always the lowest among the three size AP. For class AP, AP$_{Sc}$ lags significantly behind the other three classes because it has the smallest number of instances. In terms of efficiency, large parameters and FLOPs result in low FPS on AGX, with a maximum FPS of 7.4, which is hardly deployable on underwater robot. Finally, we also found that ResNet101 was not significantly improved over ResNet50, which means that a very deep network may not be useful for detecting small creatures in underwater scenarios. 
% Therefore, there is still large space to improve, such as overcoming the long tail distribution or proposing an efficiency/accuracy-balanced detector. 

Consequently, the design of high accuracy and high efficiency detector is still the main direction in this field and there is still large space to improve the performance.
In order to achieve this goal, a shallow backbone with strong multi-scale feature fusion ability can be proposed to extract the discriminant features of small scale aquatic organisms; a specially designed training strategy may overcome the DUO's long-tail distribution, such as a more reasonable positive/negative label sampling mechanism or a class-balanced image allocation strategy within a training batch.

\section{Conclusion}
In this paper, we introduce a dataset (DUO) and a corresponding benchmark to fill in the gaps in the community. DUO contains a variety of underwater scenes and more reasonable annotations. Benchmark includes efficiency and accuracy indicators to conduct a comprehensive evaluation of the \emph{SOTA} decoders. The two contributions could serve as a reference for academic research and industrial applications, as well as promote community development.
\bibliographystyle{IEEEbib}
\bibliography{egbib}

\end{document}